# A Comprehensive Review of Image Enhancement Techniques

Raman Maini and Himanshu Aggarwal

**Abstract:** Principle objective of Image enhancement is to process an image so that result is more suitable than original image for specific application. Digital image enhancement techniques provide a multitude of choices for improving the visual quality of images. Appropriate choice of such techniques is greatly influenced by the imaging modality, task at hand and viewing conditions. This paper will provide an overview of underlying concepts, along with algorithms commonly used for image enhancement. The paper focuses on spatial domain techniques for image enhancement, with particular reference to point processing methods and histogram processing.

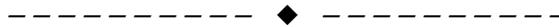

## I. Introduction

Image enhancement is basically improving the interpretability or perception of information in images for human viewers and providing `better' input for other automated image processing techniques. The principal objective of image enhancement is to modify attributes of an image to make it more suitable for a given task and a specific observer. During this process, one or more attributes of the image are modified. The choice of attributes and the way they are modified are specific to a given task. Moreover, observer-specific factors, such as the human visual system and the observer's experience, will introduce a great deal of subjectivity into the choice of image enhancement methods. There exist many techniques that can enhance a digital image without spoiling it. The enhancement methods can broadly be divided in to the following two categories:

1. Spatial Domain Methods
2. Frequency Domain Methods

In spatial domain techniques [1], we directly deal with the image pixels. The pixel values are manipulated to achieve desired enhancement. In frequency domain methods, the image is first transferred in to frequency domain. It means that, the Fourier Transform of the image is computed first. All the enhancement operations are performed on the Fourier transform of the image and then the Inverse Fourier transform is performed to get the resultant image. These enhancement operations are performed in order to modify the image brightness, contrast or the distribution of the grey levels. As a consequence the pixel value (intensities) of the output image will be modified according to the transformation function applied on the input values.

-------------------------------------------------------------------------------------------------

• Raman Maini is working as a Reader (Computer Engineering), University College of Engineering, Punjabi University, Patiala.
• Himanshu Aggarwal is working as a Reader (Computer Engineering), University College of Engineering, Punjabi University, Patiala.

Image enhancement is applied in every field where images are ought to be understood and analyzed. For example, medical image analysis, analysis of images from satellites etc.

Image enhancement simply means, transforming an image $f$ into image $g$ using $T$. (Where $T$ is the transformation. The values of pixels in images $f$ and $g$ are denoted by $r$ and $s$, respectively. As said, the pixel values $r$ and $s$ are related by the expression,

$$s = T(r) \qquad (1)$$

Where $T$ is a transformation that maps a pixel value $r$ into a pixel value $s$. The results of this transformation are mapped into the grey scale range as we are dealing here only with grey scale digital images. So, the results are mapped back into the range [0, L-1], where L=$2^k$, k being the number of bits in the image being considered. So, for instance, for an 8-bit image the range of pixel values will be [0, 255].

I will consider only gray level images. The same theory can be extended for the color images too. A digital gray image can have pixel values in the range of 0 to 255.

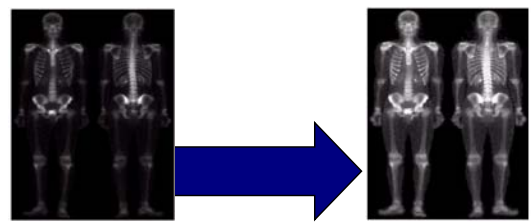

**Figure 1. Showing the effect of Image Enhancement**

Many different, often elementary and heuristic methods [2] are used to improve images in some sense. The problem is, of course, not well defined, as there is no objective measure for image quality. Here, we discuss a few recipes that have shown to be useful both for the human observer and/or for machine recognition. These methods are very problem-oriented: a method that works fine in one case may be completely inadequate for another problem. In this paper basic image enhancement techniques have been discussed with their mathmatical understanding. This paper will provide an overview of underlying concepts, along with algorithms commonly used for image enhancement. The paper focuses on spatial domain techniques for image enhancement, with particular reference to point processing methods, histogram processing.

## 2. Point Processing Operation

The simplest spatial domain operations occur when the neighbourhood is simply the pixel itself. In this case $T$ is referred to as a grey level transformation function or a point processing operation. Point processing operations take the form shown in equation (1)

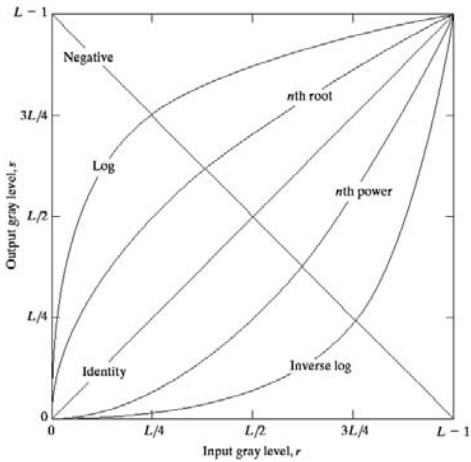

**Figure 2.** Figure shows basic grey level transformations

### 2.1 Create Negative of an Image

The most basic and simple operation in digital image processing is to compute the negative of an image. The pixel gray values are inverted to compute the negative of an image. For example, if an image of size R x C, where R represents number of rows and C represents number of columns, is represented by I(r, c). The negative N(r, c) of image I(r, c) can be computed as

N(r, c) = 255 – I(r, c) where 0 <= r <= R and 0 <= c <= C

(2)

It can be seen that every pixel value from the original image is subtracted from the 255. The resultant image becomes negative of the original image. Negative images [3] are useful for enhancing white or grey detail embedded in dark regions of an image.

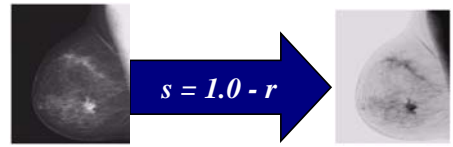

**Figure 3** Note how much clearer the tissue is in the negative image of the mammogram

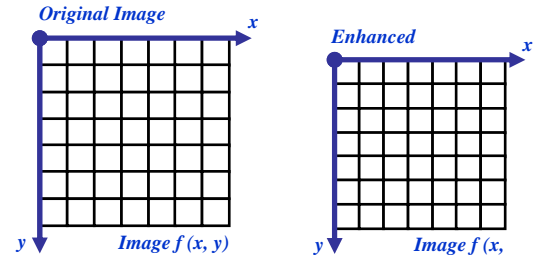

$$s = intensity_{max} - r$$

(3)

### 2.2 Thresholding Transformations

Thresholding transformations [4] are particularly useful for segmentation in which we want to isolate an object of interest from a background as shown in figure below

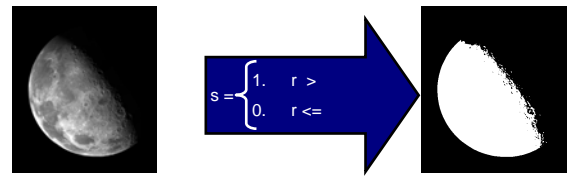

**Figure 4.** Showing effect of thresholding transformation for isolating object of interest

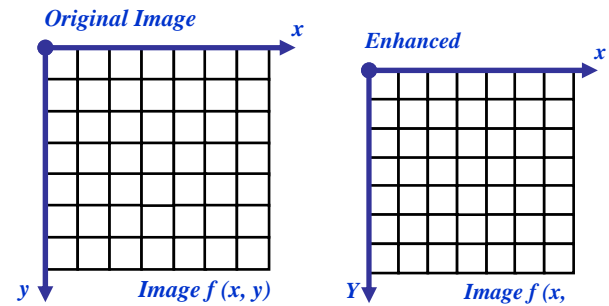

$$s = \begin{cases} 1.0 & r > threshold \\ 0.0 & r <= threshold \end{cases}$$

### 2.3 Intensity Transformation

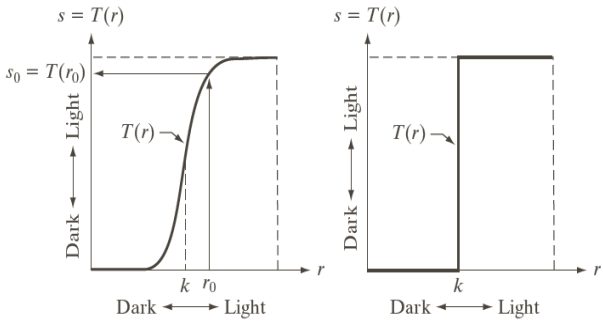

## 2.4 Logarithmic Transformations

The general form of the log transformation is

$$s = c * log \ (1 + r)$$
(4)

The log transformation maps [5] a narrow range of low input grey level values into a wider range of output values. The inverse log transformation performs the opposite transformation. Log functions are particularly useful when the input grey level values may have an extremely large range of values. In the following example the Fourier transform of an image is put through a log transform to reveal more detail

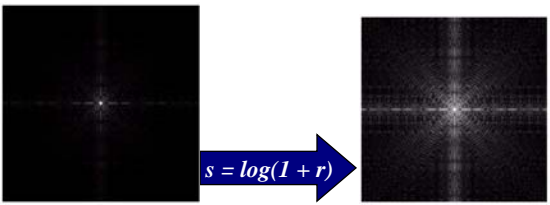

**Figure 5.** Example showing effect of Logarithmic transformation

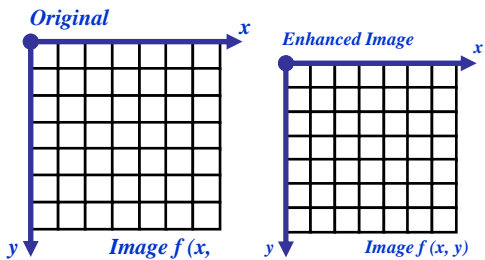

$$s = log(1 + r)$$
(5)

We usually set $c$ to 1. Grey levels must be in the range [0.0, 1.0]

## 2.5 Powers-Law Transformations

The $n$th power and $n$th root curves shown in fig. A can be given by the expression,

$$s = cr^{\gamma}$$
(6)

This transformation function is also called as *gamma* correction [6]. For various values of $\gamma$ different levels of enhancements can be obtained. This technique is quite commonly called as *Gamma Correction*. If you notice, different display monitors display images at different intensities and clarity. That means, every monitor has built-in gamma correction in it with certain gamma ranges and so a good monitor automatically corrects all the images displayed on it for the best contrast to give user the best experience. The difference between the log-transformation function and the power-law functions is that using the power-law function a family of possible transformation curves can be obtained just by varying the λ. These are the three basic image enhancement functions for grey scale images that can be applied easily for any type of image for better contrast and highlighting. Using the image negation formula given above, it is not necessary for the results to be mapped into the grey scale range [0, L-1]. Output of L-1-r automatically falls in the range of [0, L-1]. But for the Log and Power-Law transformations resulting values are often quite distinctive, depending upon control parameters like λ and logarithmic scales. So the results of these values should be mapped back to the grey scale range to get a meaningful output image. For example, Log function $s = c \ log \ (1 + r)$ results in 0 and 2.41 for $r$ varying between 0 and 255, keeping $c$=1. So, the range [0, 2.41] should be mapped to [0, L-1] for getting a meaningful image.

## OUTPUT

Enter the value for c==>1

Enter the value for gamma==>1  % for gamma value EQUALS TO 1

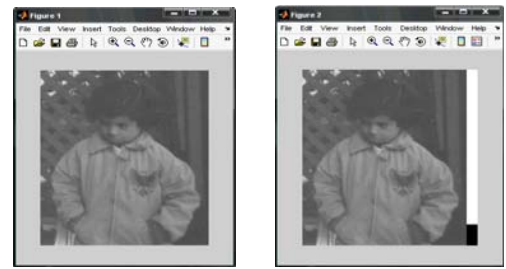

## 2.6 Piecewise Linear Transformation Functions

Rather than using a well defined mathematical function we can use arbitrary user-defined transforms.

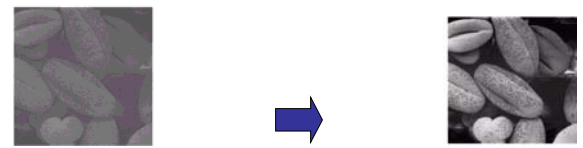

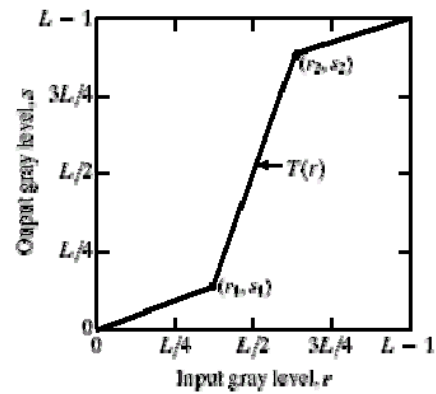

**Figure 6**. The images below show a contrast stretching linear transform to add contrast to a poor quality image

## 2.7 Grey Level Slicing

Grey level slicing [7] is the spatial domain equivalent to band-pass filtering. A grey level slicing function can either emphasize a group of intensities and diminish all others or it can emphasize a group of grey levels and leave the rest alone. Example is shown in the following figure

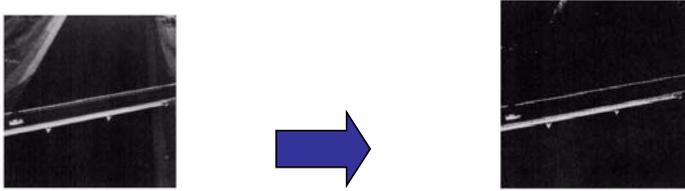

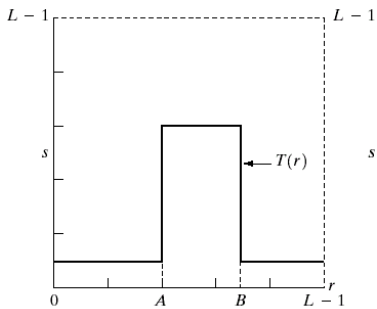

**Figure 7.** Showing example of Grey level slicing

## 3. Histogram Processing

The histogram of a digital image with intensity levels in the range [0, L-1] is a discrete function

$$h(r_k) = n_k$$

$k^{th}$ intensity value    Number of pixels in the image with intensity $r_k$

Histograms are frequently normalized by the total number of pixels in the image. Assuming an $M$ x $N$ image, a normalized histogram.

$$p(r_k) = \frac{n_k}{MN}, \quad k = 0, 1, \ldots L-1$$

is related to probability of occurrence of $r_k$ in the image

## 3.1 Histogram Equalization

Histogram equalization [8] is a common technique for enhancing the appearance of images. Suppose we have an image which is predominantly dark. Then its histogram would be skewed towards the lower end of the grey scale and all the image detail is compressed into the dark end of the histogram. If we could `stretch out' the grey levels at the dark end to produce a more uniformly distributed histogram then the image would become much clearer.

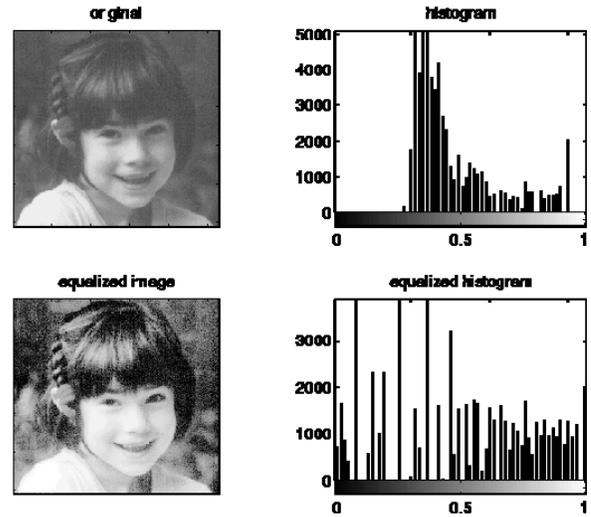

**Figure 8.** The original image and its histogram, and the equalized versions. Both images are quantized to 64 grey levels

## 3.2 Histogram Matching

Histogram equalization [9] automatically determines a transformation function seeking to produce an output image with a uniform histogram. Another method is to generate an image having a specified histogram is histogram matching.

1. Find the histogram $pr(r)$ of the input image and determine its equalization transformation

$$s = T(r) = (L-1) \int_0^r p_r(w)dw \qquad (7)$$

2. Use the specified pdf $p_z(r)$ of the output image to obtain the transformation function:

$$G(z) = (L-1) \int_0^z p_z(t)dt = s \qquad (8)$$

3. Find the inverse transformation $z = G\text{-}1(s)$ – the mapping from $s$ to $z$:

$$z = G^{-1}[T(r)] = G^{-1}(s) \qquad (9)$$

4. Obtain the output image by equalizing the input image first; then for each pixel in the equalized image, perform the inverse mapping to obtain the corresponding pixel of the output image.

Histogram matching enables us to "match" the grayscale distribution in one image to the grayscale distribution in another image

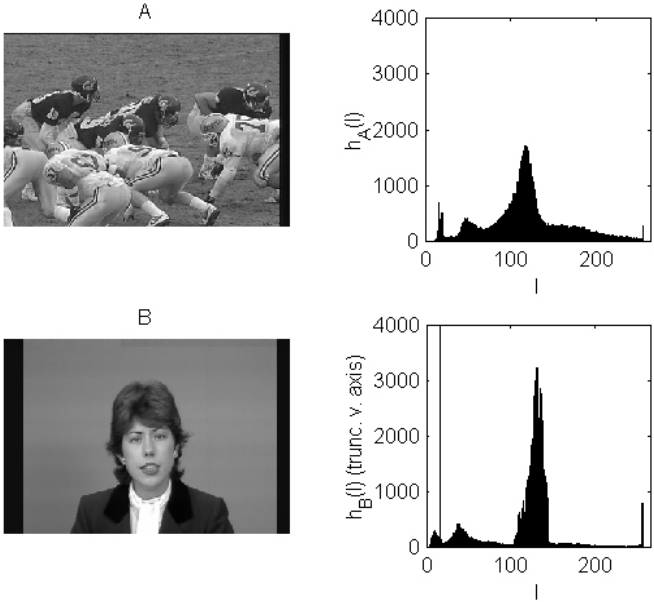

**Figure 9**. Showing histogram Matching different images

### 3.3 Local Enhancement

Previous methods of histogram equalizations and histogram matching are global. So, local enhancement [10] is used. Define square or rectangular neighborhood (mask) and move the center from pixel to pixel. For each neighborhood, calculate histogram of the points in the neighborhood. Obtain histogram equalization/specification function. Map gray level of pixel centered in neighborhood. It can use new pixel values and previous histogram to calculate next histogram.

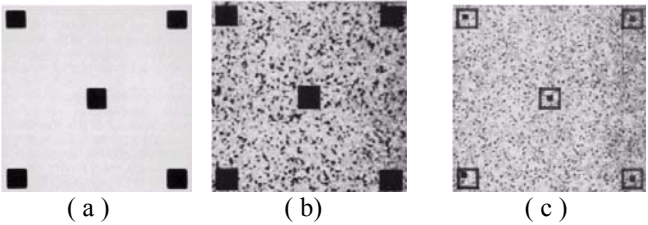

( a )  ( b )  ( c )

**Figure 10.** (a) Original Image (b) Result of global histogram Equalization (c) Result of Local histogram equalization using 7x7 neighborhood about each pixel

### 3.4 Use of Histogram Statistics for Image Enhancement

Let the intensity in an image is represented by a discrete rv $r$ in [0, L-1] and let $p$ ($r_i$) is the normalized histogram – estimate of pdf for the intensity. The nth statistical moment is

$$\mu_n(r) = \sum_{i=0}^{L-1} (r_i - m)^n p(r_i)$$

Mean value

(10)

For image intensities, a sample mean:

$$m = \frac{1}{MN} \sum_{x=0}^{M-1} \sum_{y=0}^{N-1} f(x,y)$$

(11)

and sample variance:

$$\sigma^2 = \frac{1}{MN} \sum_{x=0}^{M-1} \sum_{y=0}^{N-1} [f(x,y) - m]^2$$

(12)

As previously, we may specify global mean [11]-[12] and variance (for the entire image) and local mean and variance for a specified sub-image (subset of pixels).

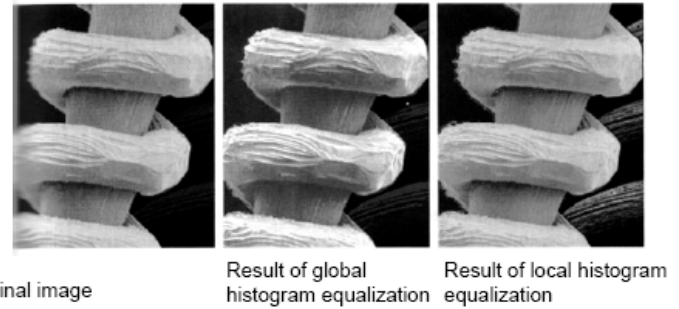

Original image    Result of global histogram equalization    Result of local histogram equalization

**Figure 11.** Showing example of using histogram statistics for image enhancement

### Concluding Remarks

Image enhancement algorithms offer a wide variety of approaches for modifying images to achieve visually acceptable images. The choice of such techniques is a function of the specific task, image content, observer characteristics, and viewing conditions. The point processing methods are most primitive, yet essential image processing operations and are used primarily for contrast enhancement. Image Negative is suited for enhancing white detail embedded in dark regions and has applications in medical imaging. Power-law transformations are useful for general-purpose contrast manipulation. For a dark image, an expansion of gray levels is accomplished using a power-law transformation with a fractional exponent. Log Transformation is Useful for enhancing details in the darker regions of the image at the expense of detail in the brighter regions the higher-level values. For an image having a washed-out appearance, a compression of gray levels is obtained using a power-law transformation with γ greater than 1. The histogram of an image (i.e., a plot of the gray-level frequencies) provides important information regarding the contrast of an image. Histogram equalization is a transformation that stretches the contrast by redistributing the gray-level values uniformly. Only the global histogram equalization can be done completely automatically

Although we did not discuss the computational cost of enhancement algorithms in this article it may play a critical role in choosing an algorithm for real-time applications. Despite the effectiveness of each of these algorithms when applied separately, in practice one has to devise a combination of such methods to achieve more effective image enhancement.